\documentclass{article}

\usepackage{arxiv}

\usepackage[utf8]{inputenc} % allow utf-8 input
\usepackage[T1]{fontenc}    % use 8-bit T1 fonts
\usepackage{hyperref}       % hyperlinks
\usepackage{url}            % simple URL typesetting
\usepackage{booktabs}       % professional-quality tables
\usepackage{amsfonts}       % blackboard math symbols
\usepackage{nicefrac}       % compact symbols for 1/2, etc.
\usepackage{microtype}      % microtypography
\usepackage{lipsum}
\usepackage{graphicx}
\usepackage{algorithm, algpseudocode}
\usepackage{amsmath,amsthm,amsfonts,amssymb,amscd, fancyhdr, color, comment, graphicx, environ}
\usepackage{tabularx}
\graphicspath{ {./images/} }

\title{Gameplay Highlights Generation}

\author{
 Vignesh Edithal \\
  \texttt{Vignesh.Edithal@amd.com} \\
   \And
 Le Zhang \\
  \texttt{Le.Zhang@amd.com} \\
  \And
 Ilia Blank \\
  \texttt{Ilia.Blank@amd.com} \\
  \And
 Imran Junejo \\
  \texttt{Imran Junejo@amd.com} \\
}

\begin{document}
\maketitle
\begin{abstract}
In this work, we enable gamers to share their gaming experience on social media by automatically generating eye-catching highlight reels from their gameplay session.
Gaming is a fast growing segment of entertainment, particularly around E-sports and Twitch.
Our automation will save time for gamers while increasing audience engagement.
We approach the highlight generation problem by first identifying intervals in the video where interesting events occur and then concatenate them.
We developed an in-house gameplay event detection dataset containing interesting events annotated by 
humans using VIA video annotator \cite{dutta2016via} \cite{dutta2019vgg}.
Traditional techniques for highlight detection such as game engine integration requires expensive collaboration with game developers.
OCR techniques which detect patches of specific images or texts require expensive per game engineering and may not generalize across game UI and different language.
We finetuned a multimodal general purpose video understanding model such as X-CLIP \cite{10.1007/978-3-031-19772-7_1} using our dataset which generalizes across multiple games in a genre without per game engineering.
Prompt engineering was performed to improve the classification performance of this multimodal model.
Our evaluation showed that such a finetuned model can detect interesting events in first person shooting games from unseen gameplay footage with more than 90\% accuracy.
Moreover, our model performed significantly better on low resource games (small dataset) when trained along with high resource games, showing signs of transfer learning.
To make the model production ready, we used ONNX libraries to enable cross platform inference.
These libraries also provide post training quantization tools to reduce model size and inference time for deployment.
ONNX runtime libraries with DirectML backend were used to perform efficient inference on Windows OS.
To conclude, we show that natural language supervision in the X-CLIP model leads to data efficient and highly performant video recognition models.
\end{abstract}

% keywords can be removed

\section{Introduction}
We tackled the problem of generating a highlight reel by first detecting time intervals in the video where an interesting event occurs and then concatenating the video clips corresponding to these events.
Detecting these time intervals is essentially modelled as a classification task where we divide a gameplay session video into \textbf{one second clips}.
Then, each clip is classified into the type of interesting event or non-interesting event (background event).
We decided to train a model to detect the \textbf{seven events} (excluding background events) across \textbf{five games} as shown in Table \ref{Dataset Stats}.
Note that not all games contain all events.
We chose games from the first person shooting (FPS) genre due to their popularity in the gaming community. 

Table \ref{Dataset Stats} lists the \textbf{five games} and \textbf{seven events} (excluding sampled background events). 

Supervised computer vision models were used to classify the video clips into one of the event types.
These models work by processing the downscaled frames from the video to extract important features which can then be used for classification.
Some models have the ability to consume information of a different modality (e.g., text, audio) to provide better generalization.
For example, feeding in the game information such as the game name can ease the task of classification.

\section{Literature Survey}

Traditional methods of event detection or action recognition use a combination of object segmentation models and OCR techniques.
For example, YOLO \cite{Redmon_2016_CVPR} is an object recognition model which can be used to draw bounding boxes around game assets and texts.
Multilingual OCR techniques can then be used to read off texts on the screen.
A deterministic algorithm can process the detected game assets and texts to predict the event type.
However, such an approach requires extensive knowledge about the game in question.
It also requires per game engineering such as asset extraction and per game algorithm.
\\
Deep Learning models for video processing were originally inspired by models for image processing.
For example, CNN networks were modified to perform 3D convolutions across time and space, such models are called 3D CNN.
C3D network \cite{Tran_2015_ICCV} has been popularly used for action recognition and video classification, it works by extracting spatio-temporal tokens from the given video input.
However, since the rise of transformer based models which led to development of large language models (LLM), the computer vision research community has been working towards replicating a similar model for images and videos.
\\
Unlike CNNs, transformers lack inherent bias for locality and translational invariance which are important for images.
However, transformers can learn these biases using attention patterns which efficiently scales to large datasets. 
This led to the invention of vision transformer (ViT) \cite{DBLP:journals/corr/abs-2010-11929} which has since revolutionized the field of computer vision.
It was able to beat the state-of-the-art at that time on the ImageNet21k dataset.
Since then, numerous image and vision models have been proposed that use ViT as a backbone for feature extraction.
ViT works by dividing an image into patches and then applies attention across those patches.
In a sense, ViT treats patches of image as words in a sentence.
\\
Video vision transformer (ViViT) \cite{Arnab_2021_ICCV} uses a factorized encoder where spatial and temporal encoders are placed in series. Spatial encoder captures latent representation per time index, then, temporal encoder models interaction between tokens at different time steps.
Even though this model has more parameters than a 3D CNN model, it requires fewer floating point operations (FLOPs) due to its efficient factorized encoder.
VideoMAE \cite{Arnab_2021_ICCV} is an autoencoder model which uses an encoder to generate a compact representation of videos and then uses a decoder to tackle tasks like predicting masked pixels.
They show that pre-training VideoMAE model on large datasets with random masked cubes (tube masking) leads to premier performance on small datasets.
It is possible to achieve high levels of masking up to 90\% -95\% masked pixels due to temporal redundancy in videos.
\\
Contrastive language image pre-training (CLIP) \cite{pmlr-v139-radford21a} is a multi-modal technique which uses natural language supervision to provide a broader and general supervision for model training.
CLIP model is trained on 400 million text image pairs sourced from the internet.
This model enables zero shot capabilities on ImageNet dataset \cite{5206848}.
A pre-trained CLIP model transfers to downstream tasks such as OCR, action recognition, geo-localization and object classification without dataset specific training.
\\
CLIP model has been extended to videos to create the X-CLIP model \cite{10.1007/978-3-031-19772-7_1} by reusing some of its weights that it learned from ImageNet.
X-CLIP introduces transformer modules such as cross frame attention, multi frame integration and video specific prompting.
The model can be seen as having two encoder modules for video and texts.
It can be used for event classification by detecting the most similar event text for the given video.
The authors of X-CLIP claim top-1 accuracy of 87.1\% on Kinetics-400 dataset \cite{DBLP:journals/corr/KayCSZHVVGBNSZ17} while using 12 times fewer FLOPs compared to Swin \cite{Liu_2021_ICCV} and ViViT models. Variations of Kinetics dataset such as Kinetics-400, are widely used to benchmark video recognition models.
\\
On the other hand, our work requires data collection and annotation of multiple representative games.
A single video recognition model can be finetuned on this data with limited transfer capabilities to other games in the same genre.
This makes our approach scalable when adding support for a large number of games.

\section{Methodologies}

\subsection{Tech Stack}

For data collection, we used the AMD Radeon ReLive software to capture gameplay videos. VGG video annotator (VIA) \cite{dutta2016via} \cite{dutta2019vgg} is an HTML based program to facilitate video annotations such as marking time intervals when interesting events occur.
Python libraries such as decord, pyav, torchvision which use ffmpeg as backend were used for frame extraction from videos. Then, the transformations made available by torchvision package were used for data pre-processing.
Python and PyTorch were used for model loading, finetuning, inference and analysis of results.
The PyTorch model was converted to open neural network exchange (ONNX) format.
This enables the consumption of the model across multiple platforms using language specific ONNX APIs.
Model inference was required to be run on Windows with AMD GPUs, which was achieved by using C++ ONNX APIs and DirectML backend. 
ONNX ecosystem also provides a suite of tools for post-training model quantization (PTQ).
Float 32 to Int 8 PTQ was done to reduce the model size by upto four times which is better suited for deployment.
We used PyTorch for ROCm which is mostly supported in a Linux environment for model training. ROCm is AMD's proprietary software for hardware acceleration of deep learning models using AMD GPUs. AMD Radeon Pro W7900 GPU was used for all experiments in this work.

\subsection{Dataset}

We created our own in-house dataset for game event detection task.
Five games mentioned in \hyperref[List of Games]{Section 2} were played for hours by testers at AMD to collect over 110 GBs worth of videos.
These videos were manually annotated by watching them and marking the time interval when one of the seven events mentioned in \hyperref[List of Events]{Section 2} occurred.
A total of \textbf{4408} individual events were annotated.
\\
The data collection process was a slow and iterative task.
We initially started with a much smaller version of the dataset and obtained the results on that dataset.
As we obtained more data from testers, we used version tagging to keep track of changes to our dataset.
Games with good initial results were the main focus for data collection.
However, this ended up resulting in better results across other games as well due to the transfer ability of our model.
The seventh and final version of our dataset contains 110 GB worth of mp4 video files arranged in five different folders per game.
The annotations were stored in json format as exported by the VIA tool.
During the data collection process it was important not to over-collect or under-collect events to prevent data skewness across game events.
For example, kill events are more common than bomb planting events in CS:GO, meaning we had to intentionally prevent annotating certain events to maintain a balanced dataset.
Since Valorant and CS:GO saw first and second best results initially, we decided to focus more on collecting events from theses games.
This is the reason behind large skewness in the number of events per games as seen in \hyperref[Table 1]{Table 1}.
\\
\begin{table}[h]
\label{Table 1}
\centering
\begin{tabular}{|c|c|c|c|c|c|}
\hline
\textbf{Event/Game} & \textbf{CS:GO} & \textbf{PUBG} & \textbf{Valorant} & \textbf{OW 2} & \textbf{Fortnite}  \\
\hline
\textbf{Kill} & 233 & 51 & 616 & 66 & 43 \\
\textbf{Death} & 241 & 29 & 595 & 38 & 10 \\
\textbf{Grenade Throw} & 271 & 21 & - & - & -  \\
\textbf{Reload} & 262 & 59 & 528 & - & 55 \\
\textbf{Bomb Planted} & 197 & - & 463 & - & - \\
\textbf{Knocked Down} & - & 31 & - & - & 18 \\
\textbf{Power Use} & - & - & 548 & 33 & - \\
\textbf{Background} & 488 & 599 & 1204 & 99 & 355 \\
\textbf{Total (excluding background)} & 1204 & 191 & 2750 & 137 & 126 \\
\textbf{Total (including background)} & 1692 & 790 & 3954 & 236 & 481 \\
\hline
\end{tabular}
\caption{Number of events per game and per event type in our dataset.
Some events do not exist in some games, for example, CS:GO does not have any power use mechanism. These missing instances are denoted by hyphen (\textbf{-}).
Background events are sampled using the algorithm mentioned in \hyperref[Algorithm 1]{Algorithm 1}. \textbf{OW 2 stands for OverWatch 2}.
}
\label{Dataset Stats}
\end{table}

\subsubsection{Sampling Background Events}
\label{Sampling Background Events}

Most of the gameplay is not interesting which can lead to a large number of false positive detections.
Therefore, it is important to finetune the model with a large number of randomly sampled events which are not interesting.
We call these events \textbf{Background Events}.
False positives mean that an interesting event (one of the seven events) is classified as a background event which can lead to a loss of an important event from the highlight clip.
It is extremely important to avoid false positives.
The model should also be resilient to false negatives to ensure that the highlights are actually interesting.
To make the model robust to background events, we sample as many background events as possible using the algorithm mentioned in \hyperref[Algorithm 1]{Algorithm 1}.
Note that due to lack of sufficient gaps and retry mechanism, it might not be possible to extract more background events.
We collect a total of \textbf{2745} background event instances across the five games to raise the number of events in our dataset to \textbf{7153}.

\begin{algorithm}
\caption{Sample Background Events}
\label{Algorithm 1}
\begin{algorithmic}[1]

\State $MAX\_RETRIES \gets 10$
\State $BUFFER\_SECS \gets 3$
\Statex

\Function{GET\_BKG\_EVENTS}{$gamename$}
    \State $bkg\_events \gets \text{dictionary}()$
    \State $filenames \gets \text{get video file names for the } gamename$
    \State $file\_event\_map \gets \text{dictionary}()$

    \For{$filename$ in $filenames$}
        \State $parsed\_events \gets \text{parse event time intervals from} filename$
        \State $file\_event\_map[filename] \gets parsed\_events$
        \State $duration \gets \text{get duration of } filename$
        \State $file\_event\_map[filename].\text{append}([[0, 0], [duration, duration]])$
        \Statex
        \Comment{Boundary Condition}
    \EndFor

    \State $retry\_counter \gets 0$
    \While{$retry\_counter < MAX\_RETRIES$}
        \State $filename \gets \text{uniform random selection of filename from } filenames$
        \State \textbf{try}
            \State \hspace{\algorithmicindent} $bk\_start\_time, bk\_end\_time \gets \text{GET\_ONE\_BKG\_EVENT}(filename, file\_event\_map)$
        \State \textbf{catch} \text{NO\_BACKGROUND\_EVENTS error}
            \State \hspace{\algorithmicindent} $retry\_counter \gets retry\_counter + 1$
        \State $file\_event\_map[filename].\text{append}([bk\_start\_time, bk\_end\_time])$
        \State $bkg\_events[filename].\text{append}([bk\_start\_time, bk\_end\_time])$
    \EndWhile

    \State \Return $bkg\_events$
\EndFunction
\Statex

\Function{GET\_ONE\_BKG\_EVENT}{$filename, file\_event\_map$}
    \State $events \gets file\_event\_map[filename]$
    \State $events \gets \text{sort events by start time}$
    \State $gaps \gets \text{list}()$

    \For{$i = 0$ to length($events$)}
        \State $gap\_start \gets i\text{th event end time} + BUFFER\_SECS$
        \State $gap\_end \gets (i + 1)\text{th event start time} - BUFFER\_SECS$
        \If{$gap\_end - gap\_start \geq 1$}
            \State $gaps.\text{append}([gap\_start, gap\_end])$
        \EndIf
    \EndFor

    \If{$gaps$ is empty list}
        \State \textbf{raise} $\text{NO\_BACKGROUND\_EVENTS error}$
    \EndIf

    \State $gap \gets \text{uniform random selection of a gap from } gaps$
    \State \Return $gap\_start\_time, gap\_end\_time$
\EndFunction

\end{algorithmic}
\end{algorithm}

\subsubsection{Data Pre-processing}

We perform a \textbf{80\% - 20\%} split of the dataset for \textbf{training} and \textbf{testing} respectively.
Common training data augmentation techniques involve randomly performing operations such as horizontal/vertical flip, colour jitter, crop, zoom.
However, these augmentations are more suitable for pre-training models (learning weights from scratch) as opposed to finetuning.
The version of \textbf{X-CLIP} we used, required \textbf{32 RGB frames} of size \textbf{224 x 224 pixels}.
We perform an uniform random sampling of frames from the one second time interval clips, for example, every fourth frame is extracted if the video has a frame rate of 128 frames per second.
Each extracted frame is then resized from their original resolution to a square image of size 224 pixels using PyTorch resize transform operation.
Since neural networks perform better on data which follows a normal distribution, we normalize the pixel values (0 to 255) of each frame using the following mean and variance values for RGB channels respectively, mean $= [123.675, 116.28, 103.55]$ and variance $= [58.395, 57.12, 57.375]$.
We also perform max scaling operation (divide by 255) on the normalized pixels to ensure all values are between 0 to 1. 

\subsection{X-CLIP Model}

Contrastive language image pre-training (CLIP) model \cite{pmlr-v139-radford21a} has demonstrated "zero-shot" generalization for various image tasks.
X-CLIP \cite{10.1007/978-3-031-19772-7_1} model expands CLIP model's ability to process videos and uses some of its modules which are trained on ImageNet instead of pre-training a model from scratch.
There are multiple variants of X-CLIP model which use a different ViT backbone and are trained on different datasets.
They also differ in terms of number of frames which can be processed which ultimately effects the FLOPs of the model.
These variants are made available as part of a open source project by Microsoft Research at \cite{xclipgithub}.
We use the X-CLIP-B/16 (which uses ViT-B backbone and patches of size 16 x 16 pixels) model which is pre-trained on Kinetics-400 dataset using a single view of 32 frames and finetuned on Kinetics-600 dataset in zero-shot fashion.
\\
X-CLIP model requires a video and a set of text prompts as input.
Video is a sequence of 32 RGB frames of size 224 x 224 pixels.
Text prompts are a list of texts where each text corresponds to an event type in the classification targets. 
At the final layer of the model, cosine similarity is calculated between the output of the video encoder and the text encoder output for each of the prompts.
Softmax function is applied to obtain per event probabilities.
Finally, the event type with the highest probability for the given one second video clip is predicted.

\subsubsection{Video Encoder}

The video encoder starts off with projecting each of the image patches into an embedding space.
These patch embeddings are concatenated, a learnable embedding $x_\textit{class}$ is prefixed and positional encoding is suffixed to create a frame embedding.

\begin{equation}
z_t^{(0)} = [x_{\textit{class}}, Px_{t,1}, ..., Px_{t,N}] + pos
\end{equation}

Here, $z_t^{(0)}$ denotes the frame embedding of frame number $t$, $P$ is the projection matrix, $x_{t,1}$ is the patch embedding of patch number 1 in frame number t, $pos$ is the postional embedding.
Each of the frame embeddings are passed through cross frame communication transformer (CCT) and the state of the learnable embedding at the output serves as the representation of the frame.
CCT models spatio-temporal information while greatly reducing computation cost as compared to factorized spatial and temporal encoders \cite{10.1007/978-3-031-19772-7_1}.
To enable cross frame information exchange, two types of attention mechanisms are used, 
cross frame fusion attention (CFA) and intra frame diffusion attention (IFA).
CFA uses a message token for each frame to abstract information exchange.
All message tokens are taken into account to learn the global spatio-temporal dependencies.

\begin{equation}
m_t^{(l)} = P^{'} x_{class}
\end{equation}

\begin{equation}
M^{'(l)} = M^{(l)} + \text{CFA}(\text{LN}(M^{(l)}))
\end{equation}

Here $m_t^{(l)}$ is the message token of frame number $t$ in layer number $l$.
$P^{'}$ is a projection of the frame representation.
LN is layer norm.
$M^{(l)}$ is the concatenation of all message tokens in layer number $l$.
\\
IFA takes the frame tokens and the associated message tokens and diffuses global spatio-temporal information into the frame tokens.

\begin{equation}
[ z^{'(l)}, m_{t}^{''(l)} ] = [ z^{(l-1)}, m_{t}^{'(l)} ] + \text{IFA}( \text{LN}( [ z_{t}^{(l - 1)}, m_t^{'(l)} ]  ) )
\end{equation}

Here, $m_t^{''(l)}$ message tokens are discarded, $z_t^{(l - 1)}$ are frame tokens from layer number $l - 1$.
Finally, a feed forward network (FFN) generates the final frame tokens for layer number $l$.

\begin{equation}
z_t^{(l)} = z_{t}^{'(l)} + \text{FFN}( \text{LN} ( z_{t}^{'(l)} ) )
\end{equation}

Alternating blocks of fusion and diffusion attention form one layer, multiple such layers are stacked to form the CCT.
The weights of the IFA are obtained from the underlying CLIP model, weights of CFA are randomly initialized.
\\
Multi frame integrator is one layer of transformer which transforms per frame representation into one video representation.

\begin{equation}
H = [h_{1}, h_{2}, ..., h_{T}]
\end{equation}

\begin{equation}
v = \text{AvgPool}(\text{MIT}(H + temp))
\end{equation}

Here, $\text{AvgPool}$ is the average pooling operation, $H$ is the concatenation of per frame output of CCT and $temp$ is the temporal position embedding.
CFA, IFA and MIT all use standard multi-head self-attention (MHSA) with feed forward networks \cite{NIPS2017_3f5ee243}.
The weights of the MIT module were randomly initialized.

\subsubsection{Text Encoder}

The pre-trained text encoder from CLIP is employed in X-CLIP.
However, a video-specific prompting module is added to enhance the text representation.
This avoids the usage of manually engineered prompts such as "A photo of a {label}", in contrast the prompt just contains the label such as "{label}".
A learnable text prompting scheme then modifies the prompt to enhance performance.
For example, video specific semantic information such as "in the water" will make it easier for the model to distinguish between "running" and "swimming" events.
Two transformer blocks consisting of MHSA and FFN are stacked in the video-specific prompting module.

\begin{align}
c^{'} &= c + \text{MHSA}(c, v) \\
c^{''} &= c^{'} + \text{FFN}(c^{'}) \\
\bar{c} &= c + \alpha c^{''}
\end{align}

Here, $c$ is the output of pre-trained text encoder, $v$ is the output of video encoder consisting of CCT and MIT, $\bar{c}$ is the final text encoding used for cosine similarity matching and $\alpha = 0.1$ is a constant.
The weights of the video-specific prompting module were randomly initialized.
It should be noted that this text encoder supports a tiny token length of 77 tokens which makes the X-CLIP model unsuitable for processing prompts larger than 77 characters.

\subsubsection{Architecture and Observations}

For a network architecture diagram of the X-CLIP model, please refer to \cite{10.1007/978-3-031-19772-7_1}.
Due to the robustness of natural language supervision, a single view inference of X-CLIP model achieves the same performance as multi-view (10 views) inference which is computationally expensive and scales linearly.
X-CLIP uses sparse sampling of frames for both training and inference to achieve the best result.
This makes X-CLIP computationally efficient as compared to ViVIT which uses dense sampling of frames and multi view inference.
Moreover, the factorized encoder of ViVIT where each frame is first encoded separately before processing by temporal encoder is unable to fully utilize the temporal cues due to its late fusion strategy.
CCT on the other hand uses early fusion technique to better model temporal dependencies.
CCT is also computationally efficient than the factorized encoder setup.

\subsection{X-CLIP Finetuning}
\label{X-CLIP finetuning}

\begin{figure}
    \centering
    \includegraphics[width=0.9\linewidth]{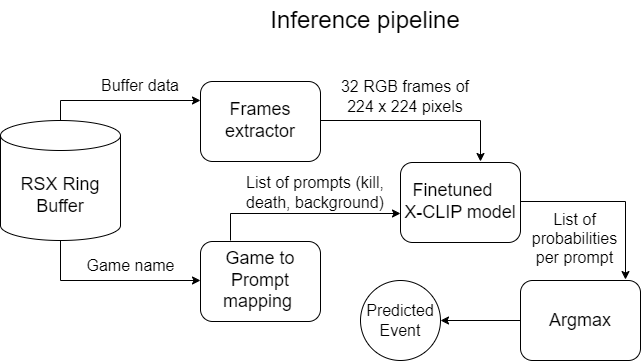}
    \caption{Pipeline block diagram representing the inference process}
    \label{fig:inference}
\end{figure}

\begin{figure}
    \centering
    \includegraphics[width=1.0\linewidth]{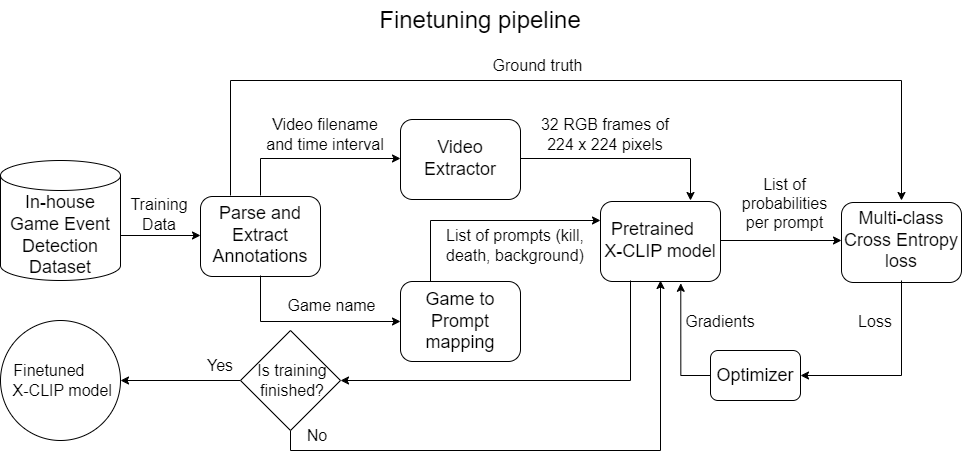}
    \caption{Pipeline block diagram representing the finetuning process}
    \label{fig:finetune}
\end{figure}

The classes in Kinetics-400 dataset on which X-CLIP was pre-trained are not related to events occurring in FPS games.
Domain difference between pre-training dataset and finetuning dataset reduces the classification performance of the model.
However, since the X-CLIP model is good at "zero-shot learning" tasks, it should be feasible to finetune the model with FPS game events.
We fix the weights of the text encoder (including the video-specific prompting module) and only update the weights of the video encoder (CCT, MIT modules).
Even though updating the weights of the text encoder leads to higher generalisation, we observed that the VRAM resources consumed are not feasible for our hardware.
In addition to updating the weights of the video encoder, finetuning also involves replacing the final fully-connected layer for classification with a new layer with the same number of neurons as the number of classes we want to predict, which is seven.
This layer is the only randomly initialized layer in the network and is kept unlocked so that it can be updated using backpropagation.
\\
We used the class labels as text prompts ("grenade throw", "power use") for the X-CLIP model.
However, the classification performance was not satisfactory and it did not show signs of knowledge transfer between games.
To tackle this, we came up with the idea of per game prompts list by following a prompting template.
Our prompt template is of the form "Game Name. Event Name. Description of the Event".
For example, to detect the event that player killed an opponent in CS:GO, we would use the following prompt "CSGO. Kill. Player in front of the gun falls down."
We observed that this kind of prompt engineering improves the classification performance of the X-CLIP model.
Theoretically speaking, injecting the game information in the prompt makes the classification task easier for the model, since it only has to distinguish between events given a game.
Moreover, not every game has every event, by avoiding unnecessary prompts for a given game (Power Use prompt for CS:GO) we prevent the model from getting confused and accidentally predicting an event which does not exist in the game.
Complete list of prompts can be found in the \hyperref[Prompts used per game]{Prompts section in appendix}.
\\
For finetuning the model, we use the AdamW \cite{abs-1711-05101} optimizer to update weights, CosineLRS \cite{LoshchilovH16a} scheduler for decaying learning rate and Cross Entropy loss function for backpropagation. We also perform hyper-parameter tuning and present our results in the \hyperref[Hyperparameters]{Hyperparameters section in appendix}.
We use a random number generator (RNG) seed to reproduce the dataset split and background event sampling.
We finetune for 10 epochs which takes around 13 hours 20 minutes to finish.
We use a batch size of 4 and number of dataloader worker processes as 4.
We save the model weights for the best validation accuracy to prevent overfitting.
\\
The pipeline for finetuning the X-CLIP model and then using the finetuned model for inference were implemented as Python notebooks.
Finetuning is performed using the 80\% split of the dataset and inference is performed using the remaining 20\% of the dataset.
The finetuning process is illustrated in Figure \ref{fig:finetune}.
The inference process is illustrated in Figure \ref{fig:inference}

\subsection{X-CLIP: Post Training Quantization}

The model size on disk was 768 MB which is too large to feasibly package within the existing software.
This size is due to the 131.5 million parameters (weights) in the video encoder and 63.5 million parameters in the text encoder, totalling to 195 million parameters for the model.
We used ONNX quantization tools which converts the model weights from 32 bit floating point (FP32) to 8 bit signed integer (INT8) data type.
This process is called post training quantization (PTQ).
This reduces the model size by four times to 192 MB with a caveat of loss in classification accuracy.
Moreover, inference speed and GPU and VRAM utilization can be dramatically reduced by performing inference of the quantized model on AI accelerators which support INT8 inference.\\
The PyTorch model is first converted into ONNX format using PyTorch APIs.
ONNX models can be consumed using a wide range of languages (C++, Java, Python) and hardware (CPU, GPU, TPU) making it feasible to deploy the model in almost any environment.
ONNX quantization tools can perform PTQ of ONNX models by inserting Quantize and DeQuantize nodes in the network such as to sandwhich the existing operator nodes (matrix multiplication).
De-Quantize nodes convert the stored INT8 weights to FP32 to be used by the operator node and the output is then converted from FP32 to INT8 using Quantize node.
However, the parameters of these nodes must be calibrated based on the input data that the model expects to see in production.
We observed that just using a single instance from testing dataset as the calibration dataset works well!
The overall classification accuracy decreased from 94.3\% to 92.9\% due to the PTQ process.

\section{Results}

A detailed evaluation of X-CLIP model is presented in this section.
During finetuning our model was able to produce \textbf{99.76\%} average accuracy on the training dataset and \textbf{94.3\%} average accuracy on testing dataset.
This shows that the model is neither over-fitting nor under-fitting, it generalizes well.
We finetune the model using \textbf{all games} in our training dataset and present the classification metrics on the testing dataset in Table \ref{Game Classification Performance Non-Quantized}. Classification performance of the quantized model is represented in Table \ref{Game Classification Performance Quantized}.

\begin{table}[h]
\centering
\begin{tabularx}{\textwidth}{|X|X|X|X|}
\hline
\textbf{Testing data} & \textbf{Accuracy (\%)} & \textbf{Avg. F-score} & \textbf{Avg. AUC score} \\
\hline
All games & 94.3 & 0.93 & 0.99 \\
CSGO & 95.0 & 0.95 & 0.99 \\
Valorant & 95.0 & 0.95 & 0.99 \\
PUBG & 90.9 & 0.84 & 0.97 \\
Fortnite & 87.6 & 0.69 & 0.89 \\
Overwatch 2 & 89.7 & 0.90 & 0.99 \\
\hline
\end{tabularx}
\caption{Per game classification performance of finetuned X-CLIP model.
This is the original \textbf{non-quantized model}. Results are averaged over all event types. AUC score is calculated using OVO setting.}
\label{Game Classification Performance Non-Quantized}
\end{table}

\begin{table}[h]
\centering
\begin{tabularx}{\textwidth}{|X|X|X|X|}
\hline
\textbf{Testing data} & \textbf{Accuracy (\%)} & \textbf{Avg. F-score} & \textbf{Avg. AUC score} \\
\hline
All games & 92.9 & 0.93 & 0.99 \\
CSGO & 92.6 & 0.92 & 0.99 \\
Valorant & 95.2 & 0.96 & 0.99 \\
PUBG & 90.3 & 0.83 & 0.97 \\
Fortnite & 85.4 & 0.67 & 0.92 \\
Overwatch 2 & 76.9 & 0.76 & 0.95 \\
\hline
\end{tabularx}
\caption{Per game classification performance of finetuned then \textbf{quantized} X-CLIP model. Results are averaged over all event types. AUC score is calculated using OVO setting.}
\label{Game Classification Performance Quantized}
\end{table}

\begin{table}[h]
\centering
\begin{tabularx}{\textwidth}{|X|X|X|}
\hline
\textbf{Event type} & \textbf{Accuracy (\%) for quantized model} & \textbf{Accuracy (\%) for non-quantized model} \\
\hline
Kill & 93.6 & 92.6 \\
Death & 96.2 & 98.9 \\
Grenade Throw & 93.1 & 93.1 \\
Power Use & 83.6 & 87.9 \\
Reload & 86.2 & 90.6 \\
Bomb planted & 99.2 & 99.2 \\
Knocked Down & 90.0 & 80.0 \\
Background & 94.2 & 95.1 \\
\hline
\end{tabularx}
\caption{Per event classification performance averaged across all games containing that event. Accuracy for quantized model is lower in all cases except the kill event.}
\label{Event Classification Performance}
\end{table}

\subsection{Classification Performance}

Classification performance per game and per event type in the testing dataset is separately calculated.
We report classification metrics such as accuracy, F-score and AUC score.
One vs one (OVO) configuration was used to calculate AUC in a multi-class setup since it is less sensitive to dataset imbalance.
These metrics have been averaged over all event types.
Classification performance of the quantized model is expected to be worse compared to the original model.
However, it can be clearly seen in Table \ref{Game Classification Performance Quantized} that the performance degradation is not significant.
Considering the low resource utilization benefits of a quantized model, we deemed such a tradeoff to be acceptable.
\\
Our model performs with greater than 90\% accuracy on some of the most popular FPS games such as CS:GO, Valorant and PUBG.
Drop in classification performance on Fortnite and OW2 can be attributed to the fact that these games are less representative in our dataset as can be seen in \hyperref[Table 1]{Table 1}.
However, we observed better performance in Fortnite and OW2 as compared to training on those games individually.
This shows the model's ability to transfer knowledge from high resource games (CSGO, Valorant, PUBG) to low resource games (Fortnite, OW2).
Quantized model performance on low resource games takes a hit due to the lack of representation of these games in the calibration dataset used to quantize the model.
\\
The ease of detecting different event types can be judged by looking at the per event type classification performance presented in Table  \ref{Event Classification Performance}.
It is evident that detecting the death event (opponent kills the player and player dies) is much easier than detecting the kill event (player kills the opponent).
This is due to the fact that many games have a clear transition scene or display a huge and easy to detect text when the player dies.
However, there is rarely a clear transition scene in FPS games when a kill occurs.
This reasoning can also be applied to the bomb planted event since this event takes considerable time to occur (for example, planting a bomb in CS:GO can take up to 5 seconds) making it easier for the model to detect.
Our model performs well in detecting background events accurately, this is important to ensure that the highlight reel is informative and interesting.
Our model is also able to detect quick and transient events such as grenade throw which can also be interspersed with other event types simultaneously (such as grenade throw and kill event).
\\
\subsection{Classification Performance On Unseen Game}
To check the performance of the model on an unseen game we downloaded a 20 minute video of \textbf{Apex Legends} gameplay from YouTube.
We chose this game due to its resemblance to top performing FPS games in our dataset such as CS:GO and Valorant.
The video was divided into one second chunks and 32 frames were extracted from each second followed by frame resizing and normalization of pixel values.
Since the model does not know what "Apex Legends" means, we decided to use "Unknown" as the game name in our prompting template.
To detect kill, death and background events we used the prompts "Unknown. Kill.". "Unknown. Death.", "Unknown. Background." respectively.
Note that we do not describe the event in our prompts.
The video was manually annotated with one second time intervals when a kill or death even occurred, this was used for performance evaluation.
\\
To give the model more context and to prevent loss of accuracy due to annotation inaccuracies we used a sliding window algorithm for event detection.
We use a \textbf{sliding window which is three seconds long}. If a kill/death is detected in any of the three 1 second clips in the sliding window, we predict that event for that window.
If a sliding window overlapped with manual annotation of the same event type, it was deemed correct.
We never encountered a case where both kill/death were predicted in a sliding window even though that might occur in actual gameplay.
\\
Sliding window based accuracy for each event type in Apex Legends is presented in Table \ref{Unseen game performance}.
Some kill events were predicted as background events and all death events were correctly predicted.
It is to be noted that kill events are more important from gameplay highlight reel perspective.
\\
\begin{table}[h]
\centering
\begin{tabularx}{\textwidth}{|X|X|X|X|X|}
\hline
\textbf{Kill} & \textbf{Death} & \textbf{Background} & \textbf{Average} \\
\hline
69.0 & 100.0 & 96.7 & 89.2 \\
\hline
\end{tabularx}
\caption{Accuracy (\%) calculated using sliding window technique across various events in \textbf{Apex Legends}. Average accuracy is also presented.}
\label{Unseen game performance}
\end{table}
\\
\subsection{Runtime Performance}
A detailed evaluation of runtime performance of the quantized finetuned X-CLIP model is presented to study its impact on gameplay performance.
We used tools such as MSI Afterburner and Riva Tuner Statistics Server on Windows OS to collect time series data of the following metrics: Average frames per second (FPS), CPU \%, GPU \%, VRAM and RAM usage.
Fortnite (with EPIC graphics settings) was used to study impact on gameplay.
Model inference was performed using a Python script.
Note that, inference of 1 second gameplay took around 0.15 seconds which meant the script had to sleep until next second of gameplay was available.
This meant model inference occurred in short bursts during the gameplay.
Another interesting observation is the ability of the model to \textbf{cache the prompt embeddings} which means that the number of prompts (or number of events detected) do not impact the time/resources taken for model inference.
Since the prompt texts never change, the text encoder is only ran once and the prompt embeddings are cached.
Resource utilization metrics were collected with both gameplay and background model inference running on the same GPU.
Upon comparison with metrics collected with just the gameplay running, we can determine the runtime performance of the model.
We observed a \textbf{13 FPS drop in gameplay performance} with average FPS of 80 with X-CLIP running in the background.
RAM utilization was 800 MB, VRAM utilization was 1500 MB.
There was no significant increase in CPU and GPU utilization percentage.
Our solution is particularly feasible for systems with an integrated GPU (iGPU) or inference processing unit (IPU) along with a dedicated GPU (dGPU).
In such a setup, model inference can be offloaded to iGPU or IPU while gameplay rendering occurs on the dGPU.
This will ensure minimum impact on gameplay performance and resource utilization.
Moreover, some IPUs support efficient inference of INT8 quantized models.

\section{Conclusions and Future Research}

To conclude, this work delivers an event detection prototype for first person shooting (FPS) games.
Many use cases can be built around detected events such as highlight generation, player ranking and gameplay suggestions.
We demonstrated that an open-source computer vision foundation model such as X-CLIP can be finetuned to obtain more than 90\% accuracy for game event detection in popular games such as CS:GO and Valorant.
It is clear that natural language supervision enables efficient and performant learning from small scale datasets.
Moreover, we showed that performing INT8 quantization of the model results in feasible model deployment in terms of resource utilization on consumer grade hardware.
AMD can integrate such a model in their software to enhance the gameplay experience for the gamers who own AMD GPUs.
\\
In terms of future research, it will be interesting to study \textbf{how X-CLIP understands gameplay videos} without any manual engineering.
For example, we could mask the pixels corresponding to in-game assets such as texts, notifications and other UI elements to see if the model is still able to detect events without any UI supervision.
Another future direction is to \textbf{apply this model to other gameplay genres} such as platforming, strategy, racing etc...
\textbf{Prompt engineering} can also be explored further.
Guidelines can be built around how to describe an event to enhance its detection accuracy.
In our work we did not use the audio modality of gameplay videos.
However, this is an important data point for event detection.
\textbf{Integrating an audio model with CLIP} based model and finetuning these networks in a joint fashion is sure to enhance classification performance.
For example, in CS:GO a unique sound is played whenever a kill event occurs, this can be used to minimize confusion in the detection of kill event.
Confusion occurs when the model detects a dead opponent and assumes that the player scored the kill even though it might have been another player who actually scored the kill.
We intend to publish our work as a research paper in external computer vision and gaming conferences.

\bibliographystyle{unsrt}  
\bibliography{template}

\newpage

\section{Appendix}

\subsection{Prompts}
\label{Prompts used per game}

We use the following list of prompts for each game using the prompting template mentioned in \hyperref[X-CLIP finetuning]{X-CLIP finetuning} section.
\begin{itemize}
    \item CSGO. kill. person in front of the gun gets shot and falls down and dies.
    \item CSGO. death. person holding the gun gets shot and falls down and dies.
    \item CSGO. grenade throw.
    \item CSGO. reload. gun is moved around to insert ammunition.
    \item CSGO. bomb planted. buttons on the bomb are being pressed
    \item CSGO. background. The gun is not firing. No person gets shot. No person dies.
    \item PUBG. kill.
    \item PUBG. death.
    \item PUBG. grenade throw.
    \item PUBG. reload.
    \item PUBG. knocked down.
    \item PUBG. background.
    \item Valorant. kill.
    \item Valorant. death.
    \item Valorant. power use.
    \item Valorant. reload.
    \item Valorant. bomb planted.
    \item Valorant. background.
    \item OW2. kill. red skull flashes in the middle side.
    \item OW2. death. grey bar flashes in the lower middle side.
    \item OW2. power use. white circle with blue boundary flashes in the lower middle side then it says zero inside the circle.
    \item OW2. background. no kill. no death. no power use.
    \item Fortnite. kill.
    \item Fortnite. death.
    \item Fortnite. reload.
    \item Fortnite. knocked down.
    \item Fortnite. background.
\end{itemize}

\subsection{Hyperparameters}
\label{Hyperparameters}

These are the hyper-parameter values obtained after manual tweaking.
We use optimizer and the learning rate (LR) scheduler from PyTorch library.
\\
AdamW optimizer uses beta1 $= 0.9$, beta2 $= 0.98$, eps = $1e-8$ and weight decay $= 8e-5$.
\\
Cosine LR Scheduler decays the learning rate every epoch in a cosine function fashion starting from LR $= 1e-3$ to a minimum LR $= 8e-7$.
\\
Training hyper-parameters include, batch size $= 4$, number of epochs $= 10$ with check-pointing on best validation accuracy.

\newpage

\end{document}